\documentclass{article} 

\usepackage{arxiv,times}

\usepackage{amsmath,amsfonts,bm}

\def\eqref#1{equation~\ref{#1}}

\def\1{\bm{1}}

\DeclareMathAlphabet{\mathsfit}{\encodingdefault}{\sfdefault}{m}{sl}
\SetMathAlphabet{\mathsfit}{bold}{\encodingdefault}{\sfdefault}{bx}{n}

\usepackage{hyperref}
\usepackage{url}
\usepackage{graphicx}
\usepackage{booktabs, nicematrix}
\usepackage{multirow}
\usepackage{xcolor}
\usepackage{colortbl}
\newcommand{\hd}[1]{\begin{tabular}[c]{@{}c@{}}#1\end{tabular}} 
\usepackage[capitalise]{cleveref}
\graphicspath{{figures/}} 

\usepackage{enumitem}
\usepackage{fvextra}
\usepackage{fancyvrb}
\usepackage{spverbatim}

\title{Write Once, Plan Many: Building Reusable Visual Planning Systems from One Scene}

\author{Ruixiao Yang$^{1}$, Yilun Hao$^{1}$, Mingxin Yu$^{1}$, Yang Zhang$^{2}$, and Chuchu Fan$^{1}$\\[0.1cm]
MIT$^{1}$, MIT-IBM Watson AI Lab$^{2}$\\[0.1cm]
\texttt{\{ruixiao, yilunhao, yumx35, chuchu\}@mit.edu}, \texttt{Yang.Zhang2@ibm.com}}

\iclrfinalcopy 

\begin{document}

\maketitle

\begin{abstract}
Vision-language models (VLMs) can reason about visual planning tasks, but querying them for every instance incurs repeated inference cost and latency. Generating reusable planning programs offers an alternative, but constructing reliable planning systems for new problem domains without extensive supervision remains challenging. We present a training-free framework that constructs reusable perception and planning programs from a domain specification and a single practice scene without a paired solution. The central challenge is obtaining useful feedback from the limited input. Our framework addresses it by leveraging VLM's ability to provide a renderer and a dynamic model based on its scene understanding. Rendering detected states enables comparison with the practice scene, while checking simulated action outcomes against the domain specification helps expose errors in rule implementation. Together, these checks guide refinement of the grounding program and planner without requiring solution labels. After refinement and component selection, the programs are frozen and solve new instances without further VLM calls. The same framework constructs systems across domains without manual pipeline redesign. Across fifteen domain variants spanning 2D and 3D scenes with discrete and continuous state, our systems achieve 99.3\% average planning success, compared with 80.9\% for the strongest domain-level program-generation baseline and 90.7\% for the strongest instance-level baseline. Cost analyses show a favorable success–cost trade-off, and ablations confirm the contributions of the executable domain model, execution feedback, and component selection.
\end{abstract}

\section{Introduction}
\begin{figure}[b]
    \centering
    \includegraphics[width=\linewidth]{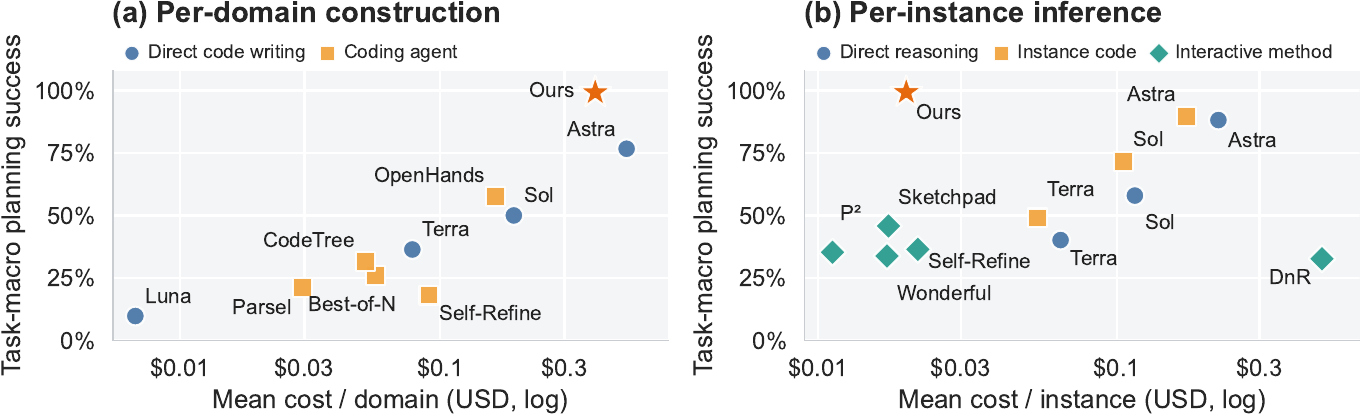}
    \caption{\textbf{Reusable planning is both capable and efficient.} Across diverse visual planning domains, our method achieves the strongest overall success–cost trade-off, outperforming domain-level program-authoring baselines while exceeding costly per-instance inference methods.}
    \label{fig:teaser-cost-perf}
\end{figure}

Vision-language models (VLMs) can interpret visual scenes in the context of natural-language goals, rules, and constraints, providing a general-purpose foundation for visual planning~\citep{nasiriany2024pivot,wake2024gpt4v,wang2024wonderful,feng2025reflective}. One way to use these capabilities is to query the VLM for each new instance, enabling scene-specific reasoning but incurring repeated inference cost and latency~\citep{hu2024sketchpad,chen2026sandboxed,hou2026codev,wu2026vtool,janjua2026perception,jeong2026draft}. When instances share the same domain rules, however, the procedure for interpreting their scenes and computing plans can potentially be reused. This motivates using the VLM to construct executable perception and planning programs that combine its broad visual and domain knowledge with efficient local execution on subsequent instances.

Our aim is to make such systems inexpensive not only to execute, but also to construct for new domains. Requiring domain-specific training data, solved demonstrations, or manually designed pipelines would introduce a new burden each time the framework is applied to a different problem domain~\citep{tang2024worldcoder,lehmann2026tycho}. We therefore study a setting in which the VLM constructs a reusable system from a domain specification, prescribed interfaces, and a single practice scene, without a paired solution or updates to model parameters. The practice scene provides a concrete observation on which to develop and execute programs, but no reference answer against which to verify their outputs. This raises a central question: \textit{How can a VLM automatically construct a reliable, reusable visual-planning system from a domain specification and one unpaired practice scene, without domain-specific training or per-instance model inference?}

The central challenge is obtaining useful correctness feedback under this limited supervision. Existing code-generation methods improve programs through execution feedback and self-generated tests~\citep{le2022coderl,ni2023lever,chen2024teaching,shinn2023reflexion,madaan2023selfrefine}, but a program can run without errors and still misread a scene or misapply a domain rule. An incorrect estimate of an object’s size or position can lead to a plan that is valid for the estimated state but infeasible in the actual scene. Similarly, an incorrect implementation of a domain rule can allow invalid actions or arrangements to pass the program’s own checks. Without a reference solution, these errors cannot be identified by simply comparing the generated plan with a known answer. Moreover, a program that succeeds on the practice scene may still fail on new instances. Reliable construction therefore requires feedback that exposes errors in both scene interpretation and rule implementation, while guiding the generation of programs that apply beyond the single available scene.

We introduce a \textbf{R}eusabe \textbf{V}isual-\textbf{P}lanning \textbf{A}uthoring (\textbf{RVPA}) framework to address this.
The key idea is to build grounding and planning programs around an executable domain model that supports both visual reconstruction and plan simulation. These capabilities turn the available scene and domain specification into complementary sources of feedback without requiring a paired solution. The model connects the components through a shared state representation and implements the visual conventions and rules common to different domain instances. During construction, the grounding program reads the practice scene, and the recovered state is rendered back into an image. Reconstruction discrepancies provide feedback for refining the grounding code. Simulated action outcomes are checked against the domain specification to refine the domain model, which then evaluates proposed plans to guide planner refinement. We instruct the VLM to author these components as reusable domain-level programs. After construction and selection, the programs are frozen and solve new instances through local execution without additional training or VLM calls. The same construction framework applies across domains without manual pipeline redesign.

We evaluate our framework across 2D and 3D visual-planning domains with discrete and continuous state, comparing against both reusable program-generation methods and per-instance VLM approaches. Across fifteen domain variants, our systems achieve 99.3\% average planning success, compared with 80.9\% for the strongest domain-level program-generation baseline and 90.7\% for the strongest instance-level baseline. Cost comparisons show a favorable success–cost trade-off, as shown in~\cref{fig:teaser-cost-perf}. Component ablations show a drop of more than 19.7\% in success rate when removing any of the executable domain model, execution feedback, and component selection, confirming their contributions.

Our contributions are threefold.
\begin{itemize}[left = 0pt, noitemsep,topsep=0pt,parsep=2pt,partopsep=2pt] 
\setlength\itemsep{0em}
    \item We present a training-free framework that uses a VLM to construct reusable visual-planning systems from a domain specification, prescribed interfaces, and a single practice scene without a paired solution. The framework supports different domains without manual pipeline redesign; each resulting system solves new instances locally without further VLM calls.

    \item We introduce a domain model that connects grounding
    and planning through a shared state representation. Rendering
    recovered states enables comparison with observations, while
    simulating actions and plans checks compliance with domain rules.
    These checks guide refinement of the domain model, grounding
    program, and planner.

    \item Across fifteen domain variants spanning 2D and 3D
    scenes with discrete and continuous state, our systems achieve
    99.3\% average planning success, versus 80.9\% for the strongest
    domain-level program-generation baseline. Cost analyses and
    component ablations examine efficiency and the sources of these gains.
\end{itemize}


\section{Related Works}

\textbf{Visual Planning with VLMs.}
Vision-language models have increasingly been used for planning from visual observations, either by generating plans directly from images or by first constructing intermediate representations that support downstream planning \citep{nasiriany2024pivot,wake2024gpt4v,huang2023voxposer,feng2025reflective}. Some approaches use VLMs to propose, refine, or verify action sequences from visual observations, including through multi-stage or multi-agent planning pipelines \citep{wang2024wonderful,yuan2024robopoint}. Others ask VLMs to formalize visually specified problems into symbolic states, planning languages, or other structured representations before invoking an external solver \citep{shirai2024vilain,dang2025planning,he2025formalize,hao2026simulation,ye2026uniplan}. These approaches span direct VLM planning, iterative visual planning, and formalization-based planning, but many either perform the visual-to-plan or visual-to-state conversion separately for each instance or rely on task-specific learned components. Our work instead studies whether a general-purpose VLM can construct a reusable visual planning pipeline in a training-free manner, so that the resulting grounding and planning components can be applied across instances.

\textbf{Programmatic and Agentic Multimodal Reasoning.}
Recent multimodal systems increasingly augment foundation models with executable code, external tools, or iterative interaction rather than relying on a single direct prediction. Code-based approaches allow models to manipulate visual inputs and inspect intermediate results through executable programs \citep{hu2024visual,chen2026sandboxed,hou2026codev}, while tool-augmented methods invoke specialized perception modules or transform their outputs into forms that are easier for language models to reason over \citep{wu2026vtool,janjua2026perception,jeong2026draft}. Related agentic frameworks further use execution feedback, self-refinement, or interaction to revise intermediate solutions and models \citep{madaan2023selfrefine,tang2024worldcoder,lehmann2026tycho}. These methods mainly use code, tools, or interaction to improve reasoning on the current instance. Our work instead uses a general-purpose VLM, without additional training, to construct reusable visual grounding and planning components around an explicit computational state, allowing later instances to be solved through execution rather than repeating the full reasoning process.

\section{RVPA Framework}
\label{sec:method}

We consider visual planning from a natural-language task specification $D$ and an observation $o$ of an instance. The observation contains the initial configuration and, when needed, a goal configuration; it may comprise several
RGB views, depth measurements, and public sensor metadata. The specification provides the task rules including planning objective, available actions, and constraints, which remain fixed across instances. Let $\Pi_D(o)$ denote the plans that satisfy these constraints and achieve the goal. Given $(D,o)$, the objective is to return
$\pi\in\Pi_D(o)$, where a plan may be an action sequence or a configuration such as a set of object placements.

\begin{figure}[t]
    \centering
    \includegraphics[width=\linewidth]{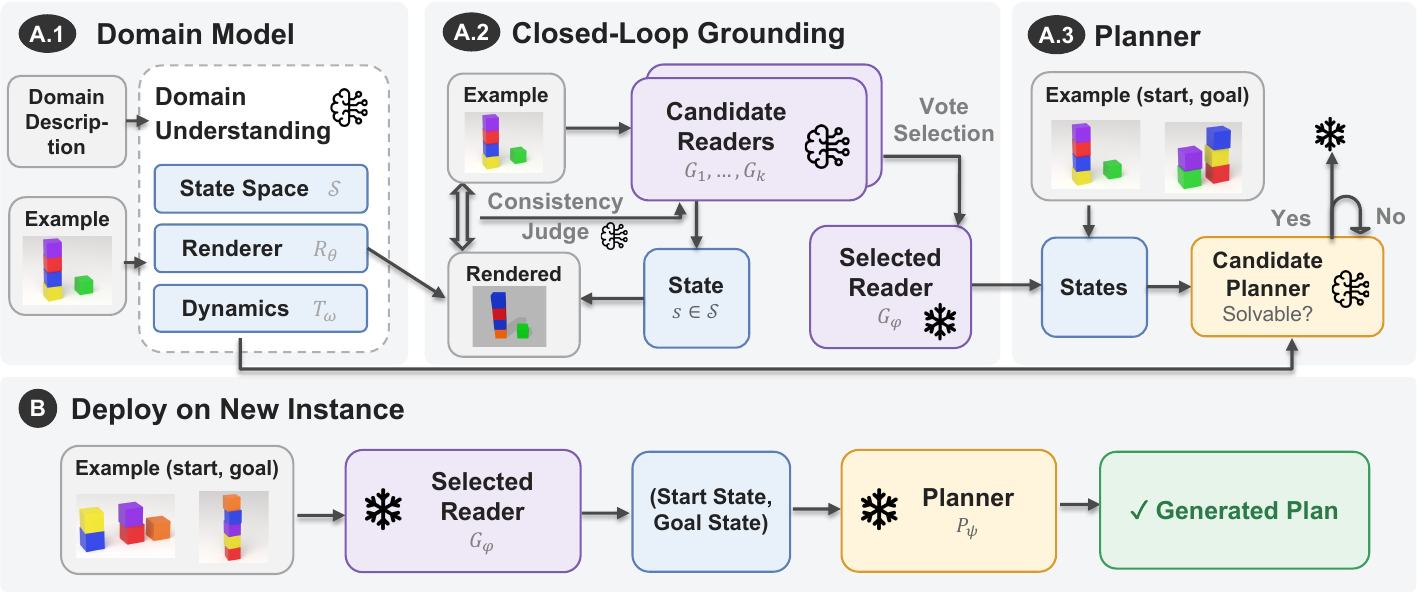}
    \caption{\textbf{Illustration of RVPA framework}. (A) Given a task description and unlabeled examples, the VLM constructs and selects reusable state-reading and planning programs using label-free feedback. (B) These programs are then frozen and applied directly to each new instance, producing plans through local execution without further VLM calls.}
    \label{fig:pipeline}
\end{figure}

As illustrated in Figure~\ref{fig:pipeline}, our approach uses a VLM to construct reusable programs once per domain, given the specification \(D\), prescribed interfaces, and a single practice observation \(o_0\). The resulting system consists of a domain model, a grounding program, and a planning program, connected through a shared state representation. The domain model includes the state representation, an appearance model that renders states into observations, and a dynamics model that simulates actions and evaluates plans. These capabilities provide reconstruction feedback for constructing the grounding program and simulation feedback for constructing the planning program. After refinement and component selection, the programs are frozen: the grounding program extracts a state from each new observation, and the planning program computes a plan without further VLM calls. 

Once construction and selection are complete, a new
instance is solved through local execution:
\begin{equation}
    s=G_\phi(o)\in\mathcal{S}, \qquad
    \pi=P_\psi(s),
    \label{eq:method-pipeline}
\end{equation}
where $\mathcal{S}$ is the shared state space, $G_\phi$ is the grounding
program, $P_\psi$ is the planning program.

\subsection{Domain Model Construction}
\label{sec:method-appearance}

We construct a domain model with state representation and appearance and dynamics components that operate on it. Given the task contract and practice images, the VLM first declares a domain-level state representation \(\mathcal{S}\) by defining a set of variables, together with an example of realization $\bar{s}$. Then, building upon the state, the VLM writes a renderer that authors an renderer model \(R_\theta\) that renders a state into a synthetic observation:
\begin{equation}
\tilde{o}=R_\theta(s),\quad s\in\mathcal{S}.
\end{equation}
The renderer candidate is then executed and refined using execution feedback on the synthetic example $\bar{s}$ provided by VLM. If the renderer has a runtime error, the VLM revises it and tests again. The selected renderer’s source code is provided to grounding-program authors so that their interpretations of coordinates, identities, and object attributes remain consistent with the model.

Similarly, based on the domain specification, prescribed interfaces, and the state representation, the VLM constructs the dynamics model $T_\omega(s,a)$, together with a set of state realizations. It simulates proposed plans and reports their validity, goal achievement, and resulting states. To check its implementation, we also execute it on its proposed synthetic states. Execution errors guide revision, and an executable candidate is retained to support planning-program construction.

\subsection{Grounding Program Construction}
The central difficulty in constructing a grounding program is the absence of a ground-truth state for the practice observation. Being able to execute alone is not a useful correctness signal: a program may run successfully while misidentifying objects, misreading geometry, or using the wrong coordinate convention. We
therefore use the renderer to obtain feedback in the observable domain.

\paragraph{Closed-loop reconstruction feedback.}
The renderer and grounding program are opposite directions of the same state representation: the grounding program maps an observation to state, whereas the renderer maps that state back to an observation. 
For a correct grounding program, their composition should preserve the task-relevant content
of the practice observation:
\begin{equation}
    o_0 \xrightarrow{\,G_k\,} s_k
    \xrightarrow{\,R_\theta\,} \tilde{o}_k,
    \qquad
    \tilde{o}_k \approx_D o_0 ,
\end{equation}
where \(\approx_D\) denotes agreement in the domain-relevant content rather than pixel-level similarity. 
The renderer is fixed before grounding-program construction, providing all candidates with the same executable definitions of object identities, coordinates, and state variables.

Given the domain specification, renderer \(R_\theta\), and practice observation
\(o_0\), the VLM independently authors \(K\) candidate grounding programs
\(\{G_k\}_{k=1}^K\). Each candidate predicts a state
\(s_k=G_k(o_0)\), which is rendered as
\(\tilde{o}_k=R_\theta(s_k)\). A separate VLM judge compares
\(\tilde{o}_k\) with \(o_0\) and identifies discrepancies in their represented
state. These discrepancies provide revision feedback to the program author;
execution errors trigger revision in the same way. The process continues until
the reconstruction is judged consistent or the revision budget is exhausted,
and the highest-scoring draft from each candidate lineage is retained.

This closed loop replaces unavailable state supervision with reconstruction supervision. An incorrect recovered state changes the rendered scene, making errors that are hidden in the latent representation observable and informative to the program author.

\paragraph{Agreement-based candidate selection.}
We maintain \(K\) independently authored grounding programs to diversify the construction process. 
After closed-loop refinement, each retained program \(G_k\) predicts a state \(s_k=G_k(o_0)\) for the practice observation. We compute pairwise agreement between these predictions and select the program
with the largest agreement neighborhood:
\begin{equation}
    k^\star =
    \arg\max_k
    \sum_{j\neq k}
    \mathbb{I}\!\left[
        \operatorname{Agree}(s_k,s_j)
    \right],
    \qquad
    G_\phi = G_{k^\star}.
\end{equation}
Ties are resolved using the prescribed selection rule. This mechanism provides a label-free selection signal: independent lineages explore alternative implementations, while agreement favors a state interpretation reproduced by multiple programs. Closed-loop feedback therefore improves each candidate individually, and cross-candidate agreement selects a stable program without access to the ground-truth state. The selected program \(G_\phi\) is frozen and reused across instances of the domain.



\subsection{Planning Program Construction}
\label{sec:method-dynamics}
Once the grounding program \(G_\phi\) is selected, construction moves from the
visual domain into the shared state space. The practice observation is mapped
to \(s_0=G_\phi(o_0)\), providing the planner with a structured description of
the scene. Grounding determines what the current configuration is, but it does
not determine which actions are valid, how they change the state, or whether
they achieve the goal. These semantics are supplied by the fixed dynamics
model \(T_\omega\).

We use \(T_\omega\) to construct a planning program \(P_\psi\) that maps
grounded states to valid, goal-reaching plans. This stage mirrors the
closed-loop construction of the grounding program: grounding uses a
\emph{render-and-compare} loop, whereas planning uses a
\emph{simulate-and-check} loop. A candidate planner proposes a plan, the
dynamics model simulates its consequences, and the resulting validity, goal
status, and diagnostic feedback guide program revision. This provides
outcome-level supervision without requiring a reference plan.

\paragraph{Simulation-guided iterative refinement.}
The VLM first authors an initial planning program \(P^{(0)}\) from the domain
specification, prescribed interfaces, and dynamics-model source. We evaluate
each successive draft on a practice set \(\mathcal{V}\) containing the recovered
practice state \(s_0\) and additional states proposed by the dynamics model. At
revision round \(t\), the planner produces a plan for each
\(s_j\in\mathcal{V}\), which is evaluated through simulation:
\begin{equation}
    \pi_j^{(t)}=P^{(t)}(s_0,s_j), \qquad
    T_\omega(s_0,\pi_j^{(t)})
    =
    \left(
        v_j^{(t)},
        g_j^{(t)},
        s_j^{\mathrm{final},(t)},
        r_j^{(t)}
    \right).
\end{equation}
Here, \(v_j^{(t)}\) indicates whether the plan is valid,
\(g_j^{(t)}\) indicates whether it achieves the goal,
\(s_j^{\mathrm{final},(t)}\) is the resulting state, and
\(r_j^{(t)}\) provides diagnostic feedback. Simulation outcomes, execution
errors, and runtime measurements are returned to the VLM, which uses them to
produce the next draft \(P^{(t+1)}\). This process evaluates whether a plan
works without requiring it to match a solution.

\paragraph{Stopping and retention.}
Each draft is scored by plan validity and goal achievement across
\(\mathcal{V}\), with partial credit for valid plans that do not achieve the
goal. Revision stops when the planner satisfies the prescribed
simulation-based acceptance criterion or when the revision budget is
exhausted. Throughout the process, we retain the highest-scoring draft; the
retained program becomes \(P_\psi\). The construction uses only feedback from
the authored dynamics model, without reference plans or benchmark correctness
feedback. The resulting planner is frozen and reused across subsequent
instances of the domain.


\subsection{Instance-Level Execution}
For each new instance, the frozen grounding program \(G_\phi\) maps the observation \(o\) to a structured state \(s\), and the frozen planning program \(P_\psi\) produces a plan \(\pi\) within the specified execution budget. Both programs run locally without further VLM calls or program revision. For evaluation, the returned plan is checked against the original instance using an independent domain validator or execution environment. Success requires satisfying the domain constraints and achieving the instance’s goal.

\section{Experiments}
\label{sec:experiments}
We evaluate whether our framework can construct reusable visual-planning systems that work for new instances without further VLM calls.
To test this capability across diverse perception and planning settings, we consider 15 task variants spanning 2D and 3D observations and discrete and continuous state. Specifically, we aim to answer these research questions:

\textbf{Q1 (Domain-level comparison):} Does our framework generate more reliable reusable code than other domain-level methods?

\textbf{Q2 (Instance-level comparison):} Can the generated code match per-instance inference while reducing recurring cost?

\subsection{Experimental Setup}

\begin{figure}[t]
    \centering
    \includegraphics[width=\linewidth]{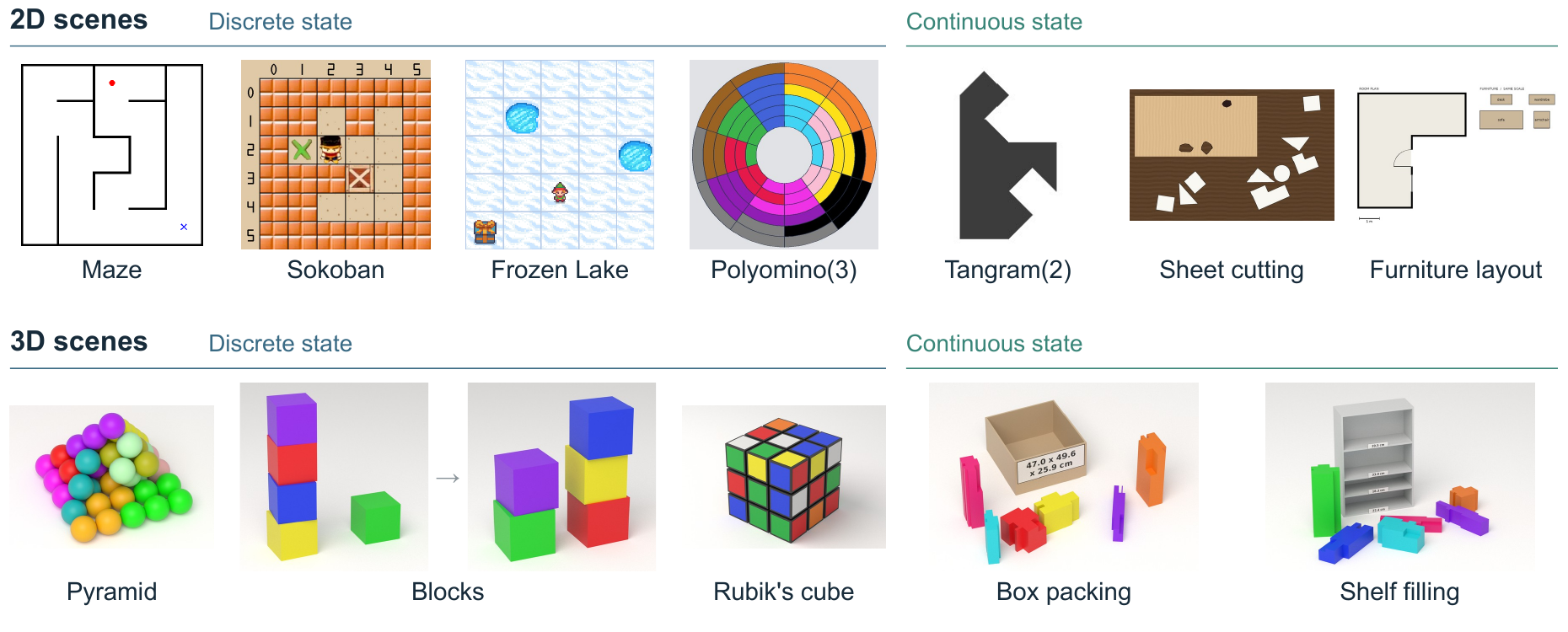}
    \caption{\textbf{Visual-planning task suite.} Representative examples summarize the 15 evaluated domain variants. The 2D domains are shown above the 3D domains, with discrete- and continuous-state domains grouped within each row. Polyomino includes original, affine, and polar variants (polar shown), while tangram includes piece and silhouette variants (silhouette shown).}
    \label{fig:tasks}
\end{figure}

\textbf{Task suite.}
We evaluate 15 visual-planning task variants spanning four settings:
2D and 3D observations with discrete and continuous oracle state, shown in~\cref{fig:tasks}. The suite covers
navigation, object rearrangement, tiling, layout optimization, packing, and
assembly. An instance provides the initial configuration and, when needed,
a goal configuration. We use RGB observations for 2D tasks and RGB-D
observations for 3D tasks. For each task, we use 20 held-out test instances. Further details are provided in~\cref{app:task-description}.

\textbf{Evaluation protocol.}
For each domain, all domain-level methods use the same fixed practice instance
and are evaluated on the same held-out instances. They receive no benchmark
correctness feedback during construction or selection, and their generated code
is frozen before held-out evaluation. Per-instance methods instead process each
held-out instance independently. 

\textbf{Metric.}
Our primary metric is success rate (SR), defined as the fraction of evaluation instances for which the returned plan satisfies all task constraints and achieves the goal. 
Success is determined by an independent task validator or execution environment. Construction failures receive zero SR, while runtime errors and timeouts count as failures on the affected instances. Because
domain-level program synthesis is stochastic, we report mean SR across independent
authoring runs.

\subsection{Domain-level Program Construction}\label{sec:exp-main}

\begin{table}[t]
\centering
\caption{\textbf{Domain-level program construction.}
Mean success rate (SR; higher is better) across 15 visual-planning domains averaged over five independent runs.
Each run constructs a fresh reusable system from the same fixed practice instance; the resulting code is frozen and evaluated on the same held-out instances. 
Our framework and coding agents use \texttt{GPT-5.6-Luna} with high reasoning effort and a 30k output-token limit, while direct code writing uses the indicated backbone among \texttt{GPT-5.6-Luna/Terra/Sol} and \texttt{GPT-6-Astra}.}

\label{tab:main-comparison}
\small
\setlength{\tabcolsep}{4.pt}
\begin{tabular}{@{}l*{10}{c}@{}}
\toprule
\multirow{2}{*}{Task} & \multicolumn{4}{c}{Direct Code Writing} & \multicolumn{5}{c}{Per-domain Coding Agent} & \multirow{2}{*}{Ours } \\
\cmidrule(lr){2-5}\cmidrule(lr){6-10}
 & Luna & Terra & Sol & Astra & \hd{Bof$N$} & \hd{Refine} & Parsel & CodeTree & OpenHands & \\
\midrule

\rowcolor{gray!20}
\multicolumn{11}{l}{\textit{\textbf{2D, discrete state}}} \\
Maze & 0.60 & 0.76 & 0.80 & 0.80 & 0.96 & 0.68 & 0.60 & 0.76 & 0.80 & \textbf{1.00} \\
Sokoban & 0.04 & 0.76 & 0.93 & \textbf{1.00} & 0.20 & 0.39 & 0.25 & 0.34 & 0.83 & \textbf{1.00} \\
Frozen Lake & 0.20 & 0.42 & 0.80 & \textbf{1.00} & 0.37 & 0.51 & 0.42 & 0.68 & \textbf{1.00} & \textbf{1.00} \\
Polyomino & 0.00 & 0.57 & 0.80 & \textbf{1.00} & 0.40 & 0.41 & 0.40 & \textbf{1.00} & \textbf{1.00} & \textbf{1.00 }\\
Polyomino (affine) & 0.19 & 0.80 & 0.80 & 0.80 & \textbf{1.00} & 0.29 & 0.60 & 0.76 & 0.69 & \textbf{1.00} \\
Polyomino (polar) & 0.38 & 0.64 & 0.38 & 0.95 & 0.20 & 0.40 & 0.00 & 0.35 & 0.60 & \textbf{1.00} \\
\midrule
\rowcolor{gray!20}
\multicolumn{11}{l}{\textit{\textbf{2D, continuous state}}} \\
Tangram (pieces) & 0.00 & 0.05 & 0.20 & 0.96 & 0.22 & 0.03 & 0.01 & 0.00 & 0.04 & \textbf{1.00} \\
Tangram (silhouette) & 0.01 & 0.04 & 0.06 & 0.68 & 0.01 & 0.00 & 0.01 & 0.00 & 0.01 & \textbf{1.00 }\\
Sheet cutting & 0.01 & 0.00 & 0.00 & 0.86 & 0.01 & 0.00 & 0.01 & 0.01 & 0.01 & \textbf{1.00} \\
Furniture layout & 0.09 & 0.55 & 0.53 & 0.37 & 0.04 & 0.05 & 0.08 & 0.16 & 0.67 & \textbf{0.89} \\
\midrule
\rowcolor{gray!20}
\multicolumn{11}{l}{\textit{\textbf{3D, discrete state}}} \\
Pyramid & 0.01 & 0.40 & \textbf{1.00} & \textbf{1.00} & 0.60 & 0.14 & \textbf{1.00} & 0.87 & 0.92 & \textbf{1.00} \\
Blocks & 0.00 & 0.19 & 0.51 & 0.61 & 0.04 & 0.00 & 0.02 & 0.04 & 0.62 & \textbf{1.00} \\
Rubik's cube & 0.00 & 0.39 & 0.40 & \textbf{1.00} & 0.00 & 0.00 & 0.00 & 0.00 & 0.74 & \textbf{1.00} \\
\midrule
\rowcolor{gray!20}
\multicolumn{11}{l}{\textit{\textbf{3D, continuous state}}} \\
Box packing & 0.00 & 0.05 & 0.67 & 0.61 & 0.00 & 0.00 & 0.00 & 0.00 & 0.46 & \textbf{1.00} \\
Shelf filling & 0.00 & 0.01 & 0.09 & 0.49 & 0.00 & 0.00 & 0.00 & 0.00 & 0.66 & \textbf{1.00} \\
\midrule
\textbf{Average} & 0.10 & 0.38 & 0.53 & 0.81 & 0.27 & 0.19 & 0.23 & 0.33 & 0.60 & \textbf{0.99}\\
\bottomrule
\end{tabular}
\end{table}

We compare our framework with methods that generate reusable visual-planning programs. This experiment tests whether our structured construction procedure produces more reliable systems than general-purpose code generation, sampling, and iterative refinement.

\paragraph{Baselines.}
We compare against two families of domain-level methods. \emph{Direct code
writing} uses a single generation using different VLMs, testing whether a stronger backbone alone produces reliable reusable code. 
We also test against Luna-based \emph{domain-level coding agent} methods: Best-of-$N$ (Bof$N$),
Self-Refine-code~\citep{madaan2023selfrefine}, Parsel~\citep{zelikman2023parsel},
CodeTree~\citep{li2025codetree}, and OpenHands~\citep{wang2024openhands}, spanning sampling and selection, iterative refinement, functional decomposition, tree-structured search, and execution-guided debugging. 

\paragraph{RQ1: Does our framework generate more reliable reusable code?}
~\cref{tab:main-comparison} shows that our framework achieves the best or
tied-best SR on all domains, with a mean SR of 99.3\%. Despite using Luna,
our method outperforms direct code writing with the substantially stronger
Astra backbone, which achieves 80.9\% mean SR. The improvement also holds when
the backbone is fixed: among the Luna-based domain-level agents, the strongest
method, OpenHands, reaches 60.3\% mean SR.

The performance gap widens in more challenging settings. Even when taking the
strongest baseline in each regime, mean SR decreases from 92.5\% on 2D
discrete-state domains to 71.8\% on 2D continuous-state domains, and from
87.0\% on 3D discrete-state domains to 56.0\% on 3D continuous-state domains.
Baseline performance also decreases from 2D to 3D for both state types. In
contrast, our framework achieves 100.0\%, 97.3\%, 100.0\%, and 100.0\% across
the four regimes, respectively. These results show that our advantage is not
limited to a particular type of domain: the generated code remains
reliable as both visual and state complexity increase.

\subsection{Comparison with per-instance inference}
\label{sec:exp-instance}
We next ask whether reusing generated code sacrifices planning performance
relative to per-instance inference, and whether avoiding repeated VLM calls
reduces cost (RQ2).

\textbf{Per-instance baselines.} We include direct reasoning and instance-specific code generation with Terra, Sol, and Astra. Visual Sketchpad~\citep{hu2024sketchpad},
P$^2$~\citep{janjua2026perception}, DnR~\citep{jeong2025draftrefine},
Self-Refine~\citep{madaan2023selfrefine}, and Wonderful
Team~\citep{wang2024wonderful} use Luna. These methods inspect each evaluation instance and may perform model reasoning, tool use, or program revision while solving it. Our method uses the same frozen systems evaluated in the preceding subsection. 

\definecolor{srbase}{RGB}{33,113,181}
\definecolor{costbase}{RGB}{217,95,2}
\newcommand{\MaxInt}{75}
\newlength{\CellW}\setlength{\CellW}{2em}
\newlength{\PairGap}\setlength{\PairGap}{1.2pt}
\newcommand{\CellFont}{\scriptsize}
\newcommand{\heatbox}[3]{%
  {\setlength{\fboxsep}{0pt}%
   \colorbox{#1!#2!white}{\makebox[\CellW]{%
     \rule[-0.45ex]{0pt}{2.1ex}\CellFont\color{black}#3}}}%
}
\newcommand{\pair}[4]{%
  $\vcenter{\offinterlineskip\vskip\PairGap
    \hbox{\heatbox{srbase}{#3}{#1}}%
    \hbox{\heatbox{costbase}{#4}{#2}}%
    \vskip\PairGap}$%
}
\newcount\TableBarLevel
\newcount\TableBarStrip
\newcommand{\tablecolorbar}[5]{
  \begingroup
  \scriptsize\setlength{\fboxsep}{0pt}\setlength{\fboxrule}{0.3pt}%
  \begin{tabular}{@{}c@{}}
    #2 \\[3pt]
    \hbox{%
      \fbox{\vbox{\offinterlineskip
        \TableBarLevel=#3\relax
        \TableBarStrip=0
        \loop
          \hbox{\color{#1!\the\TableBarLevel!white}%
            \vrule width 7pt height 2.65pt depth 0pt}%
          \advance\TableBarLevel by #4\relax
          \advance\TableBarStrip by 1
        \ifnum\TableBarStrip<76\repeat
      }}%
      \kern2pt
      \vbox to 201.4pt{#5}%
    }
  \end{tabular}%
  \endgroup
}
\newcommand{\tablevc}[1]{$\vcenter{\hbox{#1}}$}

\begin{table}[t]
\color{black}
\centering
\caption{\textbf{Frozen domain-level systems versus per-instance inference.} Each cell reports success rate (SR; \%, top; $\uparrow$ better) and API cost (cents per instance, bottom; $\downarrow$ better) over 20 held-out instances.
Direct reasoning and per-instance code use the indicated backbone; the named per-instance methods use \texttt{GPT-5.6-Luna}. Our method evaluates one frozen system per domain without runtime VLM calls; its reported cost amortizes one-time construction over the 20 instances. 
Darker blue indicates higher SR on a linear scale, and darker orange indicates lower cost on a logarithmic scale; darker cells are therefore preferred for both metrics.}
\label{tab:per-instance-comparison}
\small
\setlength{\tabcolsep}{2.5pt}
\resizebox{\linewidth}{!}{%
\tablevc{%
\begin{tabular}{@{}l*{12}{c}@{}}
\toprule
\multirow{2}{*}{Task} & \multicolumn{3}{c}{Direct reasoning} & \multicolumn{3}{c}{Per-instance code} & \multicolumn{5}{c}{Per-instance methods} & \multirow{2}{*}{Ours$^\S$} \\
\cmidrule(lr){2-4}\cmidrule(lr){5-7}\cmidrule(lr){8-12}
 & Terra & Sol & Astra & Terra & Sol & Astra & Sketchpad & P$^2$ & DnR & \hd{Self-\\Refine$^\dagger$} & \hd{Wonderful\\Team$^\dagger$} & \\
 \midrule
\rowcolor{gray!20}
\multicolumn{13}{l}{\textit{\textbf{2D, discrete state}}} \\
Maze & \pair{90.0}{0.8}{67.5}{38.8} & \pair{\textbf{100.0}}{1.7}{75.0}{33.1} & \pair{\textbf{100.0}}{2.3}{75.0}{30.7} & \pair{80.0}{1.2}{60.0}{36.3} & \pair{\textbf{100.0}}{2.4}{75.0}{30.5} & \pair{\textbf{100.0}}{2.8}{75.0}{29.2} & \pair{90.0}{\textbf{0.1}}{67.5}{54.2} & \pair{90.0}{0.3}{67.5}{47.7} & \pair{85.0}{17.9}{63.8}{14.0} & \pair{\textbf{100.0}}{0.3}{75.0}{46.7} & \pair{\textbf{100.0}}{0.6}{75.0}{41.9} & \pair{\textbf{100.0}}{0.8}{75.0}{39.6} \\
Sokoban & \pair{80.0}{0.6}{60.0}{41.2} & \pair{\textbf{100.0}}{1.4}{75.0}{34.7} & \pair{\textbf{100.0}}{2.3}{75.0}{30.5} & \pair{75.0}{0.9}{56.2}{38.8} & \pair{95.0}{3.5}{71.2}{27.2} & \pair{\textbf{100.0}}{4.0}{75.0}{26.2} & \pair{75.0}{\textbf{0.2}}{56.2}{51.7} & \pair{70.0}{0.8}{52.5}{39.4} & \pair{75.0}{25.1}{56.2}{11.3} & \pair{80.0}{0.2}{60.0}{49.6} & \pair{70.0}{0.5}{52.5}{42.5} & \pair{\textbf{100.0}}{2.0}{75.0}{31.7} \\
Frozen Lake & \pair{95.0}{0.4}{71.2}{45.0} & \pair{\textbf{100.0}}{0.8}{75.0}{39.6} & \pair{\textbf{100.0}}{1.6}{75.0}{33.9} & \pair{90.0}{0.6}{67.5}{42.0} & \pair{\textbf{100.0}}{1.4}{75.0}{35.0} & \pair{\textbf{100.0}}{1.9}{75.0}{32.4} & \pair{90.0}{\textbf{0.04}}{67.5}{63.9} & \pair{\textbf{100.0}}{0.7}{75.0}{40.8} & \pair{\textbf{100.0}}{6.2}{75.0}{22.6} & \pair{90.0}{0.1}{67.5}{54.8} & \pair{90.0}{0.3}{67.5}{47.7} & \pair{\textbf{100.0}}{1.1}{75.0}{36.8} \\
Polyomino & \pair{\textbf{100.0}}{1.4}{75.0}{34.6} & \pair{\textbf{100.0}}{4.2}{75.0}{25.9} & \pair{\textbf{100.0}}{6.2}{75.0}{22.6} & \pair{\textbf{100.0}}{1.8}{75.0}{32.5} & \pair{\textbf{100.0}}{6.2}{75.0}{22.6} & \pair{\textbf{100.0}}{7.6}{75.0}{20.9} & \pair{\textbf{100.0}}{\textbf{0.4}}{75.0}{45.4} & \pair{\textbf{100.0}}{0.9}{75.0}{38.7} & \pair{90.0}{63.4}{67.5}{3.7} & \pair{95.0}{1.3}{71.2}{35.6} & \pair{95.0}{1.9}{71.2}{32.3} & \pair{\textbf{100.0}}{1.0}{75.0}{37.9} \\
Polyomino (affine) & \pair{60.0}{11.6}{45.0}{17.5} & \pair{\textbf{100.0}}{13.9}{75.0}{16.1} & \pair{\textbf{100.0}}{8.5}{75.0}{20.1} & \pair{20.0}{6.6}{15.0}{22.1} & \pair{90.0}{13.6}{67.5}{16.2} & \pair{\textbf{100.0}}{9.6}{75.0}{19.0} & \pair{50.0}{2.2}{37.5}{31.0} & \pair{0.0}{\textbf{1.4}}{0.0}{34.8} & \pair{0.0}{83.8}{0.0}{1.4} & \pair{10.0}{5.2}{7.5}{24.1} & \pair{10.0}{3.7}{7.5}{26.8} & \pair{\textbf{100.0}}{1.8}{75.0}{32.7} \\
Polyomino (polar) & \pair{15.0}{13.9}{11.2}{16.1} & \pair{65.0}{18.6}{48.8}{13.7} & \pair{\textbf{100.0}}{9.2}{75.0}{19.4} & \pair{65.0}{7.4}{48.8}{21.2} & \pair{90.0}{12.6}{67.5}{16.9} & \pair{\textbf{100.0}}{9.9}{75.0}{18.8} & \pair{65.0}{2.8}{48.8}{29.1} & \pair{0.0}{\textbf{1.4}}{0.0}{34.5} & \pair{0.0}{81.1}{0.0}{1.7} & \pair{0.0}{4.9}{0.0}{24.6} & \pair{0.0}{3.1}{0.0}{28.4} & \pair{\textbf{100.0}}{1.6}{75.0}{33.8} \\
\midrule
\rowcolor{gray!20}
\multicolumn{13}{l}{\textit{\textbf{2D, continuous state}}}  \\
Tangram (pieces) & \pair{0.0}{13.1}{0.0}{16.6} & \pair{25.0}{21.7}{18.8}{12.5} & \pair{85.0}{50.5}{63.8}{5.6} & \pair{15.0}{8.3}{11.2}{20.3} & \pair{65.0}{19.5}{48.8}{13.3} & \pair{\textbf{100.0}}{32.9}{75.0}{9.1} & \pair{10.0}{4.4}{7.5}{25.5} & \pair{0.0}{1.2}{0.0}{36.1} & \pair{0.0}{67.6}{0.0}{3.2} & \pair{5.0}{3.8}{3.8}{26.7} & \pair{0.0}{2.3}{0.0}{30.7} & \pair{\textbf{100.0}}{\textbf{1.1}}{75.0}{37.0} \\
Tangram (silhouette) & \pair{10.0}{11.9}{7.5}{17.3} & \pair{20.0}{20.4}{15.0}{12.9} & \pair{\textbf{100.0}}{17.5}{75.0}{14.2} & \pair{25.0}{8.0}{18.8}{20.6} & \pair{70.0}{14.2}{52.5}{15.9} & \pair{95.0}{17.1}{71.2}{14.4} & \pair{10.0}{3.9}{7.5}{26.4} & \pair{0.0}{1.0}{0.0}{37.3} & \pair{0.0}{69.7}{0.0}{2.9} & \pair{0.0}{3.9}{0.0}{26.3} & \pair{0.0}{2.3}{0.0}{30.8} & \pair{\textbf{100.0}}{\textbf{0.3}}{75.0}{46.9} \\
Sheet cutting & \pair{0.0}{5.4}{0.0}{23.8} & \pair{\textbf{100.0}}{10.4}{75.0}{18.4} & \pair{\textbf{100.0}}{19.2}{75.0}{13.4} & \pair{5.0}{7.3}{3.8}{21.3} & \pair{75.0}{13.5}{56.2}{16.3} & \pair{\textbf{100.0}}{28.3}{75.0}{10.3} & \pair{15.0}{1.0}{11.2}{37.3} & \pair{15.0}{\textbf{0.7}}{11.2}{41.0} & \pair{20.0}{43.2}{15.0}{6.8} & \pair{20.0}{2.0}{15.0}{31.8} & \pair{20.0}{1.5}{15.0}{34.1} & \pair{\textbf{100.0}}{2.7}{75.0}{29.5} \\
Furniture layout & \pair{80.0}{2.1}{60.0}{31.4} & \pair{55.0}{7.7}{41.2}{20.8} & \pair{60.0}{8.8}{45.0}{19.8} & \pair{60.0}{6.1}{45.0}{22.8} & \pair{55.0}{10.8}{41.2}{18.1} & \pair{35.0}{10.6}{26.2}{18.3} & \pair{70.0}{\textbf{0.4}}{52.5}{45.4} & \pair{85.0}{0.5}{63.8}{43.2} & \pair{70.0}{29.0}{52.5}{10.1} & \pair{\textbf{90.0}}{0.7}{67.5}{40.5} & \pair{70.0}{1.5}{52.5}{34.2} & \pair{\textbf{90.0}}{3.8}{67.5}{26.6} \\
\midrule
\rowcolor{gray!20}
\multicolumn{13}{l}{\textit{\textbf{3D, discrete state}}} \\
Pyramid & \pair{10.0}{15.1}{7.5}{15.4} & \pair{25.0}{26.2}{18.8}{10.9} & \pair{80.0}{47.3}{60.0}{6.1} & \pair{95.0}{6.5}{71.2}{22.2} & \pair{\textbf{100.0}}{13.3}{75.0}{16.4} & \pair{\textbf{100.0}}{17.4}{75.0}{14.3} & \pair{60.0}{2.6}{45.0}{29.7} & \pair{5.0}{2.1}{3.8}{31.5} & \pair{0.0}{77.9}{0.0}{2.0} & \pair{10.0}{4.7}{7.5}{24.9} & \pair{5.0}{2.9}{3.8}{28.9} & \pair{\textbf{100.0}}{\textbf{0.5}}{75.0}{43.7} \\
Blocks & \pair{70.0}{0.7}{52.5}{40.6} & \pair{95.0}{1.0}{71.2}{37.5} & \pair{95.0}{1.9}{71.2}{32.2} & \pair{95.0}{0.8}{71.2}{39.4} & \pair{90.0}{1.4}{67.5}{35.0} & \pair{\textbf{100.0}}{2.2}{75.0}{31.0} & \pair{65.0}{0.3}{48.8}{47.3} & \pair{75.0}{0.5}{56.2}{43.8} & \pair{50.0}{23.9}{37.5}{11.6} & \pair{75.0}{\textbf{0.2}}{56.2}{51.2} & \pair{65.0}{0.4}{48.8}{45.0} & \pair{\textbf{100.0}}{2.4}{75.0}{30.4} \\
Rubik's cube & \pair{0.0}{6.3}{0.0}{22.5} & \pair{0.0}{16.7}{0.0}{14.6} & \pair{60.0}{81.1}{45.0}{1.7} & \pair{25.0}{8.8}{18.8}{19.8} & \pair{75.0}{14.0}{56.2}{16.0} & \pair{\textbf{100.0}}{11.6}{75.0}{17.5} & \pair{0.0}{4.5}{0.0}{25.2} & \pair{0.0}{\textbf{0.8}}{0.0}{39.2} & \pair{0.0}{47.1}{0.0}{6.1} & \pair{0.0}{2.7}{0.0}{29.3} & \pair{0.0}{1.6}{0.0}{33.7} & \pair{\textbf{100.0}}{4.5}{75.0}{25.2} \\
\midrule
\rowcolor{gray!20}
\multicolumn{13}{l}{\textit{\textbf{3D, continuous state}}} \\
Box packing & \pair{0.0}{9.5}{0.0}{19.2} & \pair{0.0}{18.2}{0.0}{13.9} & \pair{75.0}{50.1}{56.2}{5.6} & \pair{0.0}{10.2}{0.0}{18.6} & \pair{0.0}{17.9}{0.0}{14.0} & \pair{65.0}{53.0}{48.8}{5.2} & \pair{0.0}{1.8}{0.0}{32.5} & \pair{0.0}{1.2}{0.0}{36.0} & \pair{0.0}{55.9}{0.0}{4.7} & \pair{0.0}{\textbf{1.0}}{0.0}{37.5} & \pair{0.0}{1.2}{0.0}{36.0} & \pair{\textbf{100.0}}{4.1}{75.0}{26.1} \\
Shelf filling & \pair{0.0}{8.0}{0.0}{20.5} & \pair{10.0}{15.8}{7.5}{15.0} & \pair{90.0}{31.4}{67.5}{9.4} & \pair{10.0}{8.8}{7.5}{19.8} & \pair{15.0}{17.7}{11.2}{14.1} & \pair{65.0}{51.3}{48.8}{5.4} & \pair{0.0}{1.8}{0.0}{32.6} & \pair{0.0}{1.4}{0.0}{34.8} & \pair{0.0}{52.7}{0.0}{5.2} & \pair{0.0}{\textbf{1.1}}{0.0}{36.7} & \pair{0.0}{1.5}{0.0}{34.1} & \pair{\textbf{100.0}}{2.0}{75.0}{31.8} \\
\bottomrule
\end{tabular}}%
\hspace{6pt}\tablevc{\tablecolorbar{srbase}{SR (\%) $\uparrow$}{75}{-1}{%
  \hbox{\strut 100}\vfil\hbox{\strut 50}\vfil\hbox{\strut 0}}}%
\hspace{4pt}\tablevc{\tablecolorbar{costbase}{\shortstack{API cost $\downarrow$\\(cents/inst.)\\log scale}}{0}{1}{%
  \hbox{\strut 100}\vfil\hbox{\strut 10}\vfil\hbox{\strut 1}\vfil
  \hbox{\strut 0.1}\vfil\hbox{\strut 0.01}}}%
}
\par\smallskip
\end{table}

\paragraph{Cost accounting.}
For per-instance methods, we report the API cost of all model calls made while
solving an instance. For our method, runtime API cost is zero; we divide the
one-time construction cost by the 20 evaluated instances in each domain.
Construction cost includes all recorded authoring, candidate-selection,
critique, and revision calls, including unsuccessful attempts. Full accounting
details are provided in Appendix Table~\ref{tab:method-cost}.

\paragraph{RQ2: Does reuse reduce cost without sacrificing performance?}
Table~\ref{tab:per-instance-comparison} shows that our frozen systems do not
incur a performance penalty. Averaged across the 15 domains, our method
achieves 99.3\% SR, compared with 90.7\% for the strongest per-instance
baseline, code generation with \texttt{GPT-6-Astra}. Thus, despite making no VLM calls at
runtime, our generated code outperforms methods that reason or generate new code for every instance.

Our method also compares favorably in API cost. After amortizing construction
over 20 instances, it costs 1.98 cents per instance, compared with 17.35 cents
for Astra code generation and 22.53 cents for Astra reasoning, which are approximately one order of magnitude higher. 
The Luna-based per-instance methods are slightly cheaper, but their performance is substantially lower: across these methods, the highest mean SR is 46.7\%. Our method therefore provides the
strongest success--cost trade-off at the evaluated 20-instance horizon.
This advantage grows with reuse. If construction costs $C$ and the generated
system is applied to $N$ instances, its amortized API cost is $C/N$, because
execution requires no additional VLM calls. In contrast, the API cost of a
per-instance method remains approximately constant for every new instance.
The comparison therefore supports reuse as both a performance-preserving and
cost-efficient alternative to repeated inference.


\subsection{Component ablations}
\label{sec:exp-ablations}

We conduct the ablation study on 4 components of our method: the renderer, the dynamic model, the feedback revision loop for grounding, and the candidate selection. 

Table~\ref{tab:component-ablations} shows that every component contributes to overall performance, with individual ablations reducing mean success rate by 19.7--36.7 percentage points. The one-shot planner yields the largest overall decrease, underscoring the importance of iterative planning. Component contributions nevertheless vary substantially across domain types. Removing the renderer or refinement is most detrimental in 3D continuous domains, reducing success rates by 67.0 and 64.0 points, respectively. In contrast, one-shot planning incurs its largest decrease in 2D continuous domains (79.8 points). Random selection degrades performance across all domain types, including 2D discrete domains, where the other ablations retain complete success. These findings demonstrate the complementary, domain-dependent benefits of the system’s components.

\begin{table}[t]
\centering
\caption{\textbf{Component ablations across domain types.}
Each cell reports mean success rate (SR; \%) and its change
from the full system (\(\Delta\); percentage points). Results are averaged
within each domain type. Each ablation changes one component while keeping the remaining system fixed.}
\label{tab:component-ablations}
\small
\setlength{\tabcolsep}{4.3pt}
\begin{tabular}{@{}l*{5}{cc}@{}}
\toprule
& \multicolumn{2}{c}{2D Disc.} & \multicolumn{2}{c}{2D Cont.} & \multicolumn{2}{c}{3D Disc.} & \multicolumn{2}{c}{3D Cont.} & \multicolumn{2}{c}{Avg.} \\
\cmidrule(lr){2-3}\cmidrule(lr){4-5}\cmidrule(lr){6-7}\cmidrule(lr){8-9}\cmidrule(lr){10-11}
Variant & SR & $\Delta$ & SR & $\Delta$ & SR & $\Delta$ & SR & $\Delta$ & SR & $\Delta$ \\
\midrule
Full system
  & 100.0 & -- & 97.3 & -- & 100.0 & -- & 100.0 & -- & 99.3 & -- \\
w/o Renderer
  & 100.0 & 0.0 & 62.5 & $-$34.8 & 66.7 & $-$33.3 & 33.0 & $-$67.0 & 74.4 & $-$24.9 \\
w/o Grounding refinement
  & 100.0 & 0.0 & 73.8 & $-$23.5 & 66.7 & $-$33.3 & 36.0 & $-$64.0 & 77.8 & $-$21.5 \\
Random grounding selection
  & 94.3 & $-$5.7 & 78.8 & $-$18.5 & 66.3 & $-$33.7 & 57.0 & $-$43.0 & 79.6 & $-$19.7 \\
w/o planner refinement
  & 100.0 & 0.0 & 17.5 & $-$79.8 & 44.4 & $-$55.6 & 68.0 & $-$32.0 & 62.6 & $-$36.7 \\
\bottomrule
\end{tabular}
\end{table}

\section{Conclusion}
We presented a training-free framework that uses a VLM to construct reusable visual-planning systems from a domain specification, prescribed interfaces, and a single practice observation. An executable domain model connects grounding and planning through a shared state representation. Visual reconstruction provides feedback for refining grounding programs, while plan simulation guides planning-program refinement. After construction and selection, the frozen programs solve new instances through local execution without further VLM calls. Experiments across diverse domains demonstrate the potential of reusable programs to achieve high planning success while amortizing construction costs over repeated use. The same framework applies across domains, while each generated system is reused within its domain. The separation allows VLM reasoning performed during construction to support repeated visual planning without repeated model inference.

\subsection*{Reproducibility statement}
Section~3 details our program-construction framework, while Section~4
specifies the experimental setup and evaluation metrics. Appendix~B
documents the task domains, baseline configurations, experimental
protocols, and result provenance to support reproduction of the
reported results. Upon acceptance, we will publicly release our
implementation, prompts, task-generation scripts, and evaluation code
under an open-source license.

\subsection*{AI use statement}
Generative AI tools were used both as components of the research methodology
and as research-assistance tools. As part of the reported method and baselines,
vision-language and language models synthesized and revised executable code,
inferred structured scene states, evaluated visual reconstructions, and
generated plans; these uses are described in the method and experimental
sections. Generative AI tools also assisted in creating synthetic data for the
task suite, implementing and debugging research code, processing and
interpreting experimental results, creating and editing figures, surveying and
summarizing related literature, and editing the manuscript for clarity. They
were not used to formulate the research idea, hypotheses, mathematical claims,
or proofs. The initial manuscript draft was written by the authors without
generative AI assistance; for the manuscript, AI tools were used only in
subsequent editing and revision. Translation and qualitative or thematic data
analysis were not applicable to this work. The authors reviewed all AI-assisted
outputs. Synthetic data were checked against the task definitions and evaluation
protocols;
generated code was executed and checked using tests and task-specific
validators; quantitative statements were checked against experiment logs and
analysis scripts; and references and literature claims were manually verified.
The authors take responsibility for the final content of this work, including
all text, claims, data, code, figures, and other artifacts produced with the
assistance of generative AI.


\bibliography{references}
\bibliographystyle{iclr2027_conference}

\newpage
\appendix


\section{Experimental Setup}
    \subsection{Baselines Selection}
        
Our evaluation includes two groups of baselines corresponding to the two settings studied: \emph{task-level program construction} and \emph{per-instance inference}. The first group tests whether general-purpose code-generation methods can construct a reusable visual-planning system from the same task information available to our method. The second group measures the capability of methods that are allowed to inspect and reason about each evaluation instance separately. Together, these comparisons distinguish the quality of reusable system construction from the additional flexibility provided by repeated VLM inference at deployment time.

\paragraph{Task-level program construction.}
We first compare against \textbf{Direct Code Writing}, where a VLM is prompted once to construct a reusable program for the task. We evaluate this baseline with multiple backbone strengths (Luna, Terra, Sol, and Astra), which tests whether reliable reusable systems can be obtained simply by scaling the authoring model. We then compare against five coding-agent strategies using the same Luna backbone as our framework. \textbf{Best-of-N} samples multiple independent programs and selects among them; \textbf{Self-Refine-code}~\citep{madaan2023selfrefine} iteratively critiques and revises an initial program; \textbf{Parsel}~\citep{zelikman2023parsel} decomposes the target program into interacting functional components; \textbf{CodeTree}~\citep{li2025codetree} performs tree-structured search over candidate programs; and \textbf{OpenHands}~\citep{wang2024openhands} provides a general-purpose coding-agent loop with execution-guided debugging. These baselines cover major strategies for improving generated code through stronger models, sampling and selection, iterative refinement, decomposition, search, and execution feedback. All task-level methods receive the same fixed practice instance for a task, construct one reusable system, and freeze the resulting code before evaluation on new instances.

\paragraph{Per-instance inference.}
We additionally compare against methods that perform VLM inference separately for every evaluation instance. \textbf{Direct Reasoning} asks the model to produce the plan directly from the visual input, while \textbf{Instance Code} allows it to generate and execute instance-specific code before returning a solution. We evaluate both with Terra, Sol, and Astra to provide strong model-capability references. We further include representative multimodal reasoning approaches that introduce different forms of test-time computation: \textbf{Visual Sketchpad}~\citep{hu2024sketchpad} provides iterative visual reasoning through executable image-processing code; \textbf{Perception Programs (P$^2$)}~\citep{janjua2026perception} augments the VLM with structured visual cues produced by perception tools; \textbf{Draft and Refine (DnR)}~\citep{jeong2026draft} uses visual experts to refine an initial prediction; \textbf{Self-Refine}~\citep{madaan2023selfrefine} iteratively critiques and revises the current solution; and \textbf{Wonderful Team}~\citep{wang2024wonderful} uses multiple VLM roles for visual task planning and verification. These methods represent code-based, tool-augmented, refinement-based, and multi-agent approaches to improving visual reasoning at inference time.

The two groups serve different purposes. Task-level baselines directly test our central setting: whether a VLM can construct one program that remains reliable across instances. Per-instance baselines are not required to produce reusable artifacts; instead, they provide a reference for how much planning capability can be obtained when model reasoning and tool use are repeated for every new problem. We therefore report the two comparisons separately rather than treating them as a single homogeneous baseline family.  
    \subsection{Task Description}

\paragraph{Maze.}
Given an image of a maze with marked start and goal locations, find a
sequence of grid movements that reaches the goal without crossing walls.

\paragraph{Sokoban.}
Given a board containing walls, boxes, and targets, move the player to
push every box onto a target. Boxes cannot be pulled, requiring a push
sequence that avoids irreversible deadlocks.

\paragraph{Frozen Lake.}
Given a grid showing the player, holes, and a goal, find a safe route
to the goal without entering any hole. Movements are deterministic.

\paragraph{Polyomino.}
Complete a partially tiled $6 \times 10$ board using all twelve
pentominoes exactly once. Existing placements remain fixed, while the
remaining pieces may be rotated or reflected to cover every empty cell
without overlap.

\paragraph{Polyomino (affine).}
Complete the same polyomino tiling task from an image transformed by
an affine warp. The solver must recover the underlying grid and piece
placements despite the distorted geometry.

\paragraph{Polyomino (polar).}
Complete the polyomino tiling task from a board rendered in polar
coordinates. The solver must interpret the curved arrangement while
preserving the original grid-based placement constraints.


\paragraph{Tangram (pieces).}
Given images of seven scattered tangram pieces and a target silhouette,
determine how to translate, rotate, and, where permitted, flip each
observed piece to assemble the target without overlap.

\paragraph{Tangram (silhouette).}
Given a target silhouette and the seven standard tangram pieces
templates, determine a placement for every piece that covers the target
without overlap. Internal piece boundaries are not visible in the
target image.

\paragraph{Sheet cutting.}
Given a top-down image of a sheet containing unusable regions and the
part templates to be cut, arrange every part within the sheet.
Placements must avoid holes, defects, and overlap with other parts.

\paragraph{Furniture layout.}
Given a floor plan and furniture footprints drawn at the same scale,
place all furniture inside the room without overlap or obstruction of
the door swing. The remaining free space must be connected, reach the
door, and provide access to every piece.


\paragraph{Pyramid.}
Given observations of a partially completed ball pyramid, place the
remaining rigid polysphere pieces to fill all lattice sites.
Existing pieces remain fixed, and every added piece must occupy a
valid orientation without overlap.

\paragraph{Blocks.}
Given images of initial and target block configurations, produce a
sequence of moves that reconstructs the target stacks. Only clear
top blocks may be moved, and each must be placed on the table or
another clear block.

\paragraph{Rubik's cube.}
Given visual observations of a scrambled cube, recover its sticker
configuration and produce a sequence of legal face turns that restores
a uniform color on every face.


\paragraph{Box packing.}
Given multiple RGB-D views of a container and the objects to be packed,
determine object positions and orientations that fit every object
inside the container without overlap.

\paragraph{Shelf filling.}
Given RGB-D observations of the objects and the shelf, assign each object to a compartment and determine its
placement and orientation. Every object must rest on a shelf, remain
within the compartment boundaries, and avoid overlap with other objects.\label{app:task-description}

\section{Additional Result}
    \subsection{Domain-Level Run Variability}
        Table~\ref{tab:main-comparison-std} reports the variability in planning
success across independent program-construction runs. Several baselines
exhibit substantial variation, with standard deviations reaching 0.55,
indicating that construction outcomes can depend strongly on the
individual authoring run. Variability also differs across tasks: for
example, Astra is consistent on several discrete tasks but varies
considerably on furniture layout, box packing, and shelf filling.
Across the five repeated construction runs, Ours exhibits no observed
variation in success rate on 14 of the 15 tasks, with variation confined
to furniture layout (standard deviation 0.13). Together with high mean
success rates from the same runs, these results would indicate
consistent construction performance across the evaluated domains.
\begin{table}[ht]
\centering
\caption{\textbf{Domain-level program construction: standard deviations.}
Sample standard deviation of success rate across five independent runs.}
\label{tab:main-comparison-std}
\small
\setlength{\tabcolsep}{4pt}
\begin{tabular}{@{}l*{10}{c}@{}}
\toprule
\multirow{2}{*}{Task} & \multicolumn{4}{c}{Direct Code Writing} & \multicolumn{5}{c}{Per-domain Coding Agent} & \multirow{2}{*}{Ours} \\
\cmidrule(lr){2-5}\cmidrule(lr){6-10}
& Luna & Terra & Sol & Astra & \hd{Bof$N$} & \hd{Refine} & Parsel & CodeTree & OpenHands & \\
\midrule
\rowcolor{gray!20}
\multicolumn{11}{l}{\textit{\textbf{2D, discrete state}}} \\
Maze & 0.55 & 0.43 & 0.45 & 0.45 & 0.09 & 0.44 & 0.55 & 0.43 & 0.45 & 0.00 \\
Sokoban & 0.09 & 0.43 & 0.16 & 0.00 & 0.30 & 0.53 & 0.35 & 0.41 & 0.22 & 0.00 \\
Frozen Lake & 0.45 & 0.53 & 0.45 & 0.00 & 0.49 & 0.47 & 0.53 & 0.44 & 0.00 & 0.00 \\
Polyomino & 0.00 & 0.52 & 0.45 & 0.00 & 0.55 & 0.54 & 0.55 & 0.00 & 0.00 & 0.00 \\
Polyomino (affine) & 0.42 & 0.45 & 0.45 & 0.45 & 0.00 & 0.45 & 0.55 & 0.43 & 0.45 & 0.00 \\
Polyomino (polar) & 0.52 & 0.50 & 0.52 & 0.11 & 0.45 & 0.55 & 0.00 & 0.48 & 0.55 & 0.00\\
\midrule
\rowcolor{gray!20}
\multicolumn{11}{l}{\textit{\textbf{2D, continuous state}}} \\
Tangram (pieces) & 0.00 & 0.12 & 0.22 & 0.02 & 0.17 & 0.09 & 0.02 & 0.00 & 0.07 & 0.00 \\
Tangram (silhouette) & 0.02 & 0.07 & 0.09 & 0.20 & 0.02 & 0.00 & 0.02 & 0.00 & 0.02 & 0.00 \\
Sheet cutting & 0.02 & 0.00 & 0.00 & 0.10 & 0.02 & 0.00 & 0.02 & 0.02 & 0.02 & 0.00 \\
Furniture layout & 0.21 & 0.46 & 0.51 & 0.51 & 0.07 & 0.00 & 0.12 & 0.19 & 0.41 & 0.13 \\
\midrule
\rowcolor{gray!20}
\multicolumn{11}{l}{\textit{\textbf{3D, discrete state}}} \\
Pyramid & 0.02 & 0.55 & 0.00 & 0.00 & 0.55 & 0.45 & 0.00 & 0.18 & 0.19 & 0.00 \\
Blocks & 0.00 & 0.12 & 0.19 & 0.08 & 0.07 & 0.00 & 0.05 & 0.07 & 0.30 & 0.00 \\
Rubik's cube & 0.00 & 0.53 & 0.55 & 0.00 & 0.00 & 0.00 & 0.00 & 0.00 & 0.42 & 0.00 \\
\midrule
\rowcolor{gray!20}
\multicolumn{11}{l}{\textit{\textbf{3D, continuous state}}} \\
Box packing & 0.00 & 0.12 & 0.38 & 0.53 & 0.00 & 0.00 & 0.00 & 0.00 & 0.50 & 0.00 \\
Shelf filling & 0.00 & 0.02 & 0.20 & 0.50 & 0.00 & 0.00 & 0.00 & 0.00 & 0.36 & 0.00 \\
\bottomrule
\end{tabular}
\end{table}
    \subsection{Method Cost}
        \begin{table}[ht]
\centering\small
\caption{\textbf{API cost and planning success.}
Costs (USD) and success rates are averaged equally across the reported tasks.
Per-instance costs reflect inference for baselines and amortized construction for Ours.
Accounting and protocol details are provided in the text.}
\label{tab:method-cost}
\setlength{\tabcolsep}{5pt}
\begin{tabular}{@{}lrrrr@{}}
\toprule
Method & Tasks & \shortstack{\$/task\\construction} & \shortstack{\$/instance\\API cost} & Success \\
\midrule
\multicolumn{5}{@{}l}{\emph{Per-task program construction}} \\
One-shot Luna & 16 & 0.0067 & 0$^{\ast}$ & 0.10 \\
One-shot Terra & 16 & 0.0781 & 0$^{\ast}$ & 0.36 \\
One-shot Sol & 16 & 0.1917 & 0$^{\ast}$ & 0.50 \\
One-shot Astra & 16 & 0.5204 & 0$^{\ast}$ & 0.77 \\
Best-of-N & 16 & 0.0564 & 0$^{\ast}$ & 0.26 \\
Self-Refine & 16 & 0.0904 & 0$^{\ast}$ & 0.18 \\
Parsel & 16 & 0.0294 & 0$^{\ast}$ & 0.21 \\
CodeTree & 16 & 0.0515$^{\S}$ & 0$^{\ast}$ & 0.32 \\
OpenHands & 16 & 0.1634$^{\S}$ & 0$^{\ast}$ & 0.58 \\
Ours & 15 & 0.3943 & 0$^{\ast}$ & 0.99 \\
\midrule
\multicolumn{5}{@{}l}{\emph{Per-instance evaluation}} \\
Reasoning (Terra) & 16 & -- & 0.0648 & 0.40 \\
Reasoning (Sol) & 16 & -- & 0.1149 & 0.58 \\
Reasoning (Astra) & 16 & -- & 0.2188 & 0.88 \\
Instance code (Terra) & 16 & -- & 0.0541 & 0.49 \\
Instance code (Sol) & 16 & -- & 0.1051 & 0.72 \\
Instance code (Astra) & 16 & -- & 0.1706 & 0.90 \\
Sketchpad & 16 & -- & 0.0172 & 0.46 \\
P$^2$ & 16 & -- & 0.0112 & 0.35 \\
DnR & 16 & -- & 0.4861 & 0.33 \\
Self-Refine$^{\dagger}$ & 16 & -- & 0.0216 & 0.36 \\
Wonderful Team$^{\dagger}$ & 16 & -- & 0.0170 & 0.34 \\
Ours (amortized)$^{\ddagger}$ & 15 & -- & 0.0197 & 0.99 \\
\bottomrule
\end{tabular}
\end{table}

Table~\ref{tab:method-cost} separates upfront program construction from per-instance API expenditure. Ours achieves approximately 99\% task-macro success at an average construction cost of \$0.3943 per task, equivalent to \$0.0197 per instance over 20 instances. In comparison, Astra per-instance code achieves 90\% success at \$0.1706 per instance, while Astra reasoning achieves 88\% at \$0.2188. Once constructed, our programs require no further API calls, so their amortized API cost decreases with reuse. These comparisons should be interpreted with differences in task coverage, evaluation protocols, and reference-planner use in mind.

\section{Prompt Templates and a Recorded Blocks Run}
\label{app:prompts-blocks}

\subsection{General prompts for component construction}
\label{app:general-prompts}

This section gives the prompt templates used to construct the appearance
model (renderer), grounding program (perceiver), dynamics model, and planner.
It also includes the two judging prompts and the revision instructions that
connect program generation to execution feedback. The construction order is
renderer, perceiver, dynamics model, and planner. Once selected, the programs
are reused through local execution without further language-model queries.

The task specification and interfaces are supplied inputs. They define the
state fields, observation modalities, coordinate conventions, action format,
rules, and goal condition. There is no separate model call that invents the
state schema or supplies a correct terminal state. The perceiver reads the
initial configuration and, where required, a separate goal observation; the
dynamics model computes the terminal state by executing the planner's output.

For readability, the templates below replace task-dependent material with
named fields in braces. These fields are expanded before a request is sent;
they are not additional instructions to the model. The inserted renderer and
dynamics sources are complete programs. Optional observation metadata,
legends, cardinality statements, and tool notices appear only when declared
by the task contract. The templates preserve the implementation's operative
instructions, with line wrapping and task-specific signatures abstracted.
The accompanying records retain the fully instantiated Blocks prompts.

\subsubsection{Appearance model / renderer}
\label{app:prompt-renderer}

Inputs: the supplied state schema, public observation metadata, declared
rendering interface, and fixed practice observation. The output is executable
rendering code. For an RGB-D task, the interface requests both color and depth
so that subsequent reconstruction checks use corresponding modalities.

\noindent\textbf{Renderer prompt.}
\begingroup
\footnotesize
\begin{spverbatim}
{state_schema}

{optional_legend_and_observation_metadata}

Execution policy: language models may only author or revise code. Your generated functions must execute locally without calling an LLM, VLM, or remote model API, including during smoke tests and practice evaluation.

Write `{render_signature}` -> a PIL.Image that REPRODUCES the observation as closely as you can (same geometry, cell layout, element appearances, boundary lines, background). 

Reason briefly, then ONE ```python``` block defining {render_signature} (self-contained; may import PIL, numpy, cv2, math; {optional_legend_loading_instruction} compute the rest).
\end{spverbatim}
\endgroup

The RGB-D version expands \texttt{\{render\_\allowbreak{}signature\}} to include the additional
\texttt{render\_\allowbreak{}rgbd} interface. Execution failures and available rendering diagnostics
are returned through the revision prompt in Section~\ref{app:prompts-revision}. In the native Blocks
run below, renderer smoke checks used a known practice state. Observation-based
perceiver feedback should therefore not be interpreted as a claim that the
entire renderer-construction process was free of reference-state access.

\subsubsection{Grounding program / perceiver}
\label{app:prompt-perceiver}

Inputs: the state schema, observation description, selected renderer source,
fixed practice observation, and declared runtime interface. The renderer
communicates the encoding and appearance conventions; the program must
recover the actual instance from its observations.

\noindent\textbf{Perceiver prompt.}
\begingroup
\footnotesize
\begin{spverbatim}
{state_schema}

{optional_metadata}

{metric_description}

We have a TWIN renderer that APPROXIMATELY reproduces observations like this one (it is our model of the world, fit to the observation -- its exact pixel placement may be slightly off):

```python
{renderer_source}
```

Use it to understand the ENCODING -- the geometry type, the element appearances, the layout -- and the State type you must return. But do NOT compute positions purely from the twin's parameters (they are only
approximate, and the error grows across the observation). Instead, locate the state's elements and read their values, recovering the real geometry from the observation's own pixels.

You may also USE the twin inside your code: re-render your candidate reading with it and compare against the input observation at runtime, to check your own output -- and if the comparison shows your reading disagrees with the observation, reconsider your method, not only its parameters.

Your code must contain NO hardcoded pixel coordinates, angles, radii, rotations, cell sizes, or any other number read off the attached images;
compute every geometric number from the image at runtime. Choose whatever image-analysis method best fits this observation's geometry.

Execution policy: 
language models may only author or revise code. Your generated functions must execute locally without calling an LLM, VLM, or remote model API, including during smoke tests and practice evaluation.

Write `{detect_signature}` -> the state.

Installed packages you may use 
(import any of them, however you see fit):
  - open_clip: CLIP image and text encoders (zero-shot image classification)
  - transformers: pretrained vision models (DINOv2, SAM, GroundingDINO, OWL-ViT, ...)
  - torch: tensor and GPU computation
  - PIL: image loading and drawing
  - numpy: array math
  - scikit-image: classical image processing
  - cv2 (OpenCV): 
  classical computer vision (contours, Hough transforms,
    perspective and polar warps, template matching, ...)
    
Compute everything from the pixels.

First reason step by step: how is the observation laid out and how will you locate each element; how will you decide each element's value; which methods and packages will you use, and why. Then write the code.

Reply format: your step-by-step reasoning, then ONE ```python``` code block with the complete {detect_signature}. No other prose.
\end{spverbatim}
\endgroup

\subsubsection{Separate visual judge for perception}
\label{app:prompt-perception-judge}

For an eligible retained draft, the judge receives the state description and
one observed/reconstructed RGB comparison per view. Its task is to assess
state agreement while allowing stylistic differences. RGB-D reconstruction
residuals can guide the program author, but this visual judge sees RGB only.

\noindent\textbf{Perception judge prompt.}
\begingroup
\footnotesize
\begin{spverbatim}
You are checking the output of a perception program.

Each attached image (one per camera or view) has two halves labelled "observation | render of your reading": the left half is the observation from that camera, and the right half is a simulator's rendering, from the same camera, of the state that the program read. Some observations include depth; only the color images are shown here. The simulator draws in its own style: line thickness, shading, lighting, perspective details, textures and background may differ from the observation, and those differences are NOT errors. 

Decide only whether the STATE drawn on the right is the same as the state shown on the left: every element the state describes must have the same identity or color, position, size and orientation.

What the state describes:
{state_schema}

How the observation is laid out:
{layout_note}

Compare the two halves of every attached image carefully, element by element, before deciding. For each difference, say exactly which element it is, where it is, and what the observation shows versus the render. 

Reply with one JSON object only: {"verdict": "match" or "mismatch", "differences": ["one specific difference per entry"]} Use an empty differences list for a match.
\end{spverbatim}
\endgroup

A match can stop revision of that reader; a mismatch supplies specific
discrepancies for the next revision. The reconstruction objective retains
drafts, and agreement among readers selects the final reader. A judge verdict
does not itself select a reader or certify the recovered state.

\subsubsection{Dynamics model}
\label{app:prompt-dynamics}

Inputs: the task rules, state and plan formats, frozen renderer source, and
practice state produced by the selected perceiver. The author implements
plan simulation, short rule-testing demonstrations, and optional synthetic
practice variants. The renderer can then display the simulated outcomes.

\noindent\textbf{Dynamics prompt.}
\begingroup
\footnotesize
\begin{spverbatim}
Write the task's dynamics twin: an executable model of the rules, in the state format below. Implement exactly:

def simulate(state, plan):
    # -> {"valid": bool, "goal_reached": bool, "reason": str, "final_state": <state>,
    #     "tolerance": float or None, "scale": float or None}
def demo_plans(state):
    # -> list of 2-4 short plans that exercise the rules, at least one of them illegal
def variants(state, seed):
    # -> list of up to 3 states of the same format reachable from state by legal
    # actions (random, controlled by seed); return [] if the task has no actions
    # that change the initial state.
simulate executes the plan step by step under the rules; valid is False at the first illegal step (final_state = the state just before it); goal_reached says whether the final state satisfies the goal. final_state uses the SAME format as the input state so that the perception twin below can draw it. Do not mutate the input. variants are used as extra practice inputs for the solver.
The state was measured, so its values can be slightly wrong. For a valid plan, tolerance is the smallest change to any measured quantity of the state (in the state's own units) that would make this plan invalid, and scale is the largest measured extent of the state in the same units. 
Both are None when the state has no measured continuous quantities, or when the plan is invalid.

PERCEPTION TWIN (draws a state; read it to learn exactly what each field means)
```python
{twin_source}
```
The state is given as JSON in the format below (dict keys that were coordinate tuples appear as "row,col" strings). Use only Python's standard library, numpy, scipy, ortools and shapely. Use no repository imports, files, network or ground truth. Return concise reasoning followed by ONE complete fenced python block.

STATE FORMAT
{state_schema}

TASK RULES
{rules}

PLAN FORMAT
{plan_format}

PRACTICE STATE (read from a real observation by the perception program)
{practice}{practice_note}
\end{spverbatim}
\endgroup

\subsubsection{Separate judge for dynamics demonstrations}
\label{app:prompt-dynamics-judge}

The demonstrations are executed locally before judgment. The judge receives
the task rules, demonstration plans, reported outcomes, a rendered initial
state, and rendered final states for demonstrations declared valid. The
fraction of demonstrations judged correct provides construction feedback.

\noindent\textbf{Dynamics judge prompt.}
\begingroup
\footnotesize
\begin{spverbatim}
You are checking an executable model of a task's rules.

TASK RULES
{rules}

PLAN FORMAT
{plan_format}

The first image is a drawing of the initial state. For each demo below, the model executed the plan from that initial state; for demos the model called valid, the listed image is a drawing of the model's final state. Drawings come from a simulator with its own style; judge only the state they show. For each demo decide whether the model's outcome (valid or not, the reason, goal reached or not, and the final state drawn) is exactly what the rules say executing that plan produces.

DEMOS
{demos}

Reply with one JSON object only:
{"demos": [{"index": <i>, "correct": true or false, "problem": "<what is wrong, specific>"}, ...]}
\end{spverbatim}
\endgroup

These are judgments of demonstrations proposed by the dynamics author.
Independent evaluation of the final plan remains separate from this feedback.

\subsubsection{Planner}
\label{app:prompt-planner}

Inputs: the task rules, state and plan formats, selected dynamics source,
and the first practice state. Execution feedback evaluates the program on
the practice reading and up to three accepted synthetic variants. The
output is a reusable solve function rather than a solution to just the
displayed example.

\noindent\textbf{Planner prompt.}
\begingroup
\footnotesize
\begin{spverbatim}
Write a solver for the task below. Implement exactly:
def solve(state, time_limit_s=20, seed=0):
    # -> a plan in the PLAN FORMAT, or None if none is found in time
It must work on new states of this format; 
do not hard-code the practice state.
Honor time_limit_s (wall clock). 
Your plan will be executed by the dynamics twin below; a plan is correct when that twin reports valid and goal_reached. You may copy its logic.

Each call to solve runs in a fresh Python process: nothing is kept between calls, so any table or precomputation is rebuilt on every call and counts against time_limit_s.
A plan is correct when the dynamics twin reports valid and goal_reached, whatever its length, unless the task rules require otherwise. Your solver is checked on several practice states, and the measured time of each call is reported back to you. The plan is executed with the true sizes, with no tolerance, so measurement errors can cause overlaps. The dynamics twin also reports each plan's tolerance: how much the measured values could be wrong before the plan becomes invalid. Among correct plans a larger tolerance is better, and your solver is scored on it (relative to the reported scale).

DYNAMICS TWIN
```python
{dynamics_source}
```
The state is given as JSON in the format below (dict keys that were coordinate tuples appear as "row,col" strings). 
Use only Python's standard library, numpy, scipy, ortools and shapely. 
Use no repository imports, files, network or ground truth. 
Return concise reasoning followed by ONE complete fenced python block.

STATE FORMAT
{state_schema}

TASK RULES
{rules}

PLAN FORMAT
{plan_format}

PRACTICE STATE (read from a real observation by the perception program)
{practice}{practice_note}
\end{spverbatim}
\endgroup

The shared tolerance wording also appears in the recorded Blocks prompt,
although this task has no measured continuous quantities in its symbolic
state. When multiple practice states are used, \texttt{\{practice\_\allowbreak{}note\}} states their
number and clarifies that only the first appears in the prompt. The other
states contribute execution feedback. Independent authors receive an
additional author-index line.

\subsubsection{Execution feedback and revision}
\label{app:prompts-revision}

The same bounded propose-execute-revise structure is used across components.
The original authoring prompt is repeated, followed by retained code and
feedback from its execution. For the perceiver, feedback may additionally
include reconstructed views, spatial RGB/depth residuals, readback
differences, and accumulated visual-judge discrepancies. Dynamics feedback
contains the demonstration outcomes and judge critiques. Planner feedback
contains per-practice-state validity, goal achievement, elapsed time, and a
plan preview.

\noindent\textbf{Revision prompt suffix.}
\begingroup
\footnotesize
\begin{verbatim}

Your previous code:
```python
{retained_program_source}
```

It did not pass:
{execution_error_or_scored_feedback}

{optional_perception_judge_feedback}

Revise and resend the COMPLETE code.
\end{verbatim}
\endgroup

When perception discrepancies are available,
\texttt{\{optional\_\allowbreak{}perception\_\allowbreak{}judge\_\allowbreak{}feedback\}} expands to:

"An independent check compared the observation with the simulator's
rendering of what your code read, and found these differences:
\texttt{\{differences\_\allowbreak{}grouped\_\allowbreak{}by\_\allowbreak{}judged\_\allowbreak{}draft\}}"

The authoring loop retains the highest-scoring executable draft, preferring
the latest draft on a tie. Candidate selection follows the component-specific
criterion: reader agreement for perception, demonstration judgment for
dynamics, and practice execution for planning. No extra model prompt is
needed to perform these programmatic selection steps.

\subsection{Case study: a recorded Blocks construction and execution}
\label{app:blocks-case}

We illustrate the complete sequence using \texttt{blocks\_\allowbreak{}000}, the fixed practice
instance from the recorded simulated RGB-D Blocks experiment.

\subsubsection{Task specification and observation interface}
\label{app:blocks-specification}

The task is to rearrange colored blocks from an observed initial arrangement
into an arrangement shown in a separate goal observation.

Both observations use a calibrated front RGB-D camera. The color image is
640 x 480 pixels; the aligned float32 depth array measures distance along
the optical axis in centimeters. Public calibration comprises pinhole
intrinsics K = [fx, fy, cx, cy] and a 4 x 4 camera-to-world transform
\texttt{T\_\allowbreak{}world\_\allowbreak{}cam}. The world frame has z pointing upward and the table at z = 0;
camera coordinates follow the OpenCV convention. The synthetic observations
have no added depth noise. During perception authoring, the displayed
practice image places RGB above a grayscale depth visualization. The same
frozen reader is later applied to the separate goal observation.

The action rules supplied to the dynamics and planner authors are:

"A move takes the block currently on top of one stack (nothing rests on it)
and places it either on the table (starting a new stack) or on top of the
block currently on top of another stack; a covered block cannot be moved,
cannot be a destination, and a block cannot be moved onto itself."

The goal is equality of the current and target stack arrangements, ignoring
the order of stacks. A plan is a list of dictionaries:

\begingroup
\footnotesize
\begin{verbatim}
{"block": <color of moved block>,
 "onto": <color of destination block, or "table">}
\end{verbatim}
\endgroup

This is a symbolic manipulation task: successful execution requires legal
stack operations and the correct final stacking relations. The recorded
experiment does not evaluate robot trajectories or physical grasp execution.

\subsubsection{Shared state, target state, and reached final state}
\label{app:blocks-state}

The perceiver returns a list of stacks. Before planning, the current-scene
reading and goal-scene reading are combined into a dictionary. The saved
practice input to dynamics and planner construction is exactly:

\begingroup
\footnotesize
\begin{verbatim}
{
  "state": [["yellow", "blue", "red", "purple"], ["green"]],
  "goal":  [["red", "yellow", "blue"], ["green", "purple"]]
}
\end{verbatim}
\endgroup

Thus, the initial scene has a four-block tower, ordered yellow, blue, red,
purple from bottom to top, and a separate green block. The target has a
red-yellow-blue tower and a green-purple tower. Although orange belongs to
the public color vocabulary, this instance contains no orange block.

Executing the saved six-move plan with the selected dynamics program gives:

\begingroup
\footnotesize
\begin{verbatim}
{
  "valid": true,
  "goal_reached": true,
  "reason": "plan executed legally",
  "final_state": {
    "state": [["green", "purple"], ["red", "yellow", "blue"]],
    "goal":  [["red", "yellow", "blue"], ["green", "purple"]]
  },
  "tolerance": null,
  "scale": null
}
\end{verbatim}
\endgroup

The two lists inside \texttt{final\_\allowbreak{}state} differ only in stack order, so they denote
the same arrangement. This \texttt{final\_\allowbreak{}state} is a computed execution result. It
is distinct from the target description stored in goal. The step-by-step
execution producing this result appears in Section~\ref{app:blocks-planner}.

\subsubsection{Renderer: turning a symbolic state into RGB-D}
\label{app:blocks-renderer}

The selected appearance model defines:

\begingroup
\footnotesize
\begin{verbatim}
render(state, cameras) -> PIL.Image of size 640 x 960
render_rgbd(state, cameras) -> (color, depth)
\end{verbatim}
\endgroup

The first function produces the stacked RGB/depth display. The second
returns color and depth dictionaries indexed by camera name. To visualize
a planning state, the wrapper passes only its "state" field to the renderer;
the renderer itself consumes a list of stacks.

The authored renderer canonicalizes stack order by decreasing height and
then color sequence. It places axis-aligned cubes in a row, casts calibrated
camera rays against the cubes and the table plane, and uses the closest hit
to obtain optical-axis depth and visible surfaces. RGB rendering adds face
shading and shadows. The following is an exact excerpt of its geometry
construction; other rendering helpers are omitted here:

\begingroup
\footnotesize
\begin{verbatim}
def _make_blocks(stacks):
    # Cubes are 10 cm on a side.  Stacks are arranged on a centered row.
    side = 10.0
    spacing = 22.0
    n = len(stacks)
    blocks = []

    for si, stack in enumerate(stacks):
        x = (si - 0.5 * (n - 1)) * spacing
        y = 0.0
        for zi, color in enumerate(stack):
            lo = np.array([x - side / 2.0,
                           y - side / 2.0,
                           zi * side], dtype=np.float64)
            hi = np.array([x + side / 2.0,
                           y + side / 2.0,
                           (zi + 1) * side], dtype=np.float64)
            blocks.append((lo, hi, color))
    return blocks
\end{verbatim}
\endgroup

The fixed 10 cm cube size and 22 cm stack spacing belong to this authored
appearance approximation. They are not measurements returned by the
perceiver. The observed practice scene uses 5 cm cubes; consequently, the
authored rendering is too large under the native camera and crops the
purple block at the top of the tower. The archived observation/render panel
preserves this discrepancy. It illustrates why an executable renderer can
still provide imperfect visual feedback. The case study uses the original
renderer and camera parameters throughout.

\subsubsection{Perceiver: recovering stacks from color and depth}
\label{app:blocks-perceiver}

The selected reader implements:

\begingroup
\footnotesize
\begin{verbatim}
detect_state(rgb, depth, cameras) -> list[list[str]]
\end{verbatim}
\endgroup

It back-projects depth pixels into world coordinates, classifies saturated
pixels by normalized RGB distance to the authored color palette, and finds
connected components for each color. Robust world-space extents provide
component locations and an estimated cube size. The reader merges nearby
fragments of the same color, groups components by their horizontal footprint,
and sorts each group by world height to obtain a bottom-to-top stack.

The back-projection uses the supplied camera parameters. Its core is:

\begingroup
\footnotesize
\begin{verbatim}
yy, xx = np.indices((H, W), dtype=np.float64)
zc = dep
xc = (xx + 0.5 - cx) * zc / fx
yc = (yy + 0.5 - cy) * zc / fy
cam_pts = np.stack((xc, yc, zc), axis=-1)

R = T[:3, :3]
t = T[:3, 3]
world_pts = cam_pts @ R.T + t
\end{verbatim}
\endgroup

The saved output for the initial observation is
[["yellow", "blue", "red", "purple"], ["green"]]. Applying the same reader
to the goal observation produces
[["red", "yellow", "blue"], ["green", "purple"]]. Both readings were
reproduced by local execution of the frozen program when preparing this
appendix and agree with the reference stacking relations up to stack order.

Reader 0's first draft had reconstruction retention score 0.900519. Its
observed-versus-rendered normalized RGB error was 0.207300, and its relative
depth error was 0.190625. Equal modality weighting gives reconstruction
agreement 1 - (0.207300 + 0.190625)/2 = 0.801037; combining this with valid
structured output gives 0.5 + 0.5 x 0.801037 = 0.900519. The separate visual
judge returned \{"verdict":"match","differences":[]\}, stopping this author's
revision. Agreement-based selection subsequently chose reader 0 from the
five candidates.

The archived synthetic readback nevertheless contains discrepancies, and
the RGB comparison visibly crops the top block. Thus, this particular judge
verdict is not evidence of a faithful reconstruction. The reader's correctness
on the observed scene is established separately by the recorded state
evaluation. The program succeeds here by measuring the supplied RGB-D input
rather than adopting the renderer's assumed cube scale.

\subsubsection{Dynamics model: legal moves, demonstrations, and feedback}
\label{app:blocks-dynamics}

The selected program implements simulate(state, plan), \texttt{demo\_\allowbreak{}plans}(state),
and variants(state, seed). Simulation copies the current stacks, applies
one move at a time, and rejects the first move whose source or destination
is not exposed, or that places a block onto itself. A table move creates a
new singleton stack. A move onto another block removes the source's top
block and appends it to the destination stack. Goal checking canonicalizes
the stack list:

\begingroup
\footnotesize
\begin{verbatim}
def _signature(stacks):
    # Stack order is irrelevant, but duplicate stacks remain distinct.
    return tuple(sorted(tuple(stack) for stack in stacks))
\end{verbatim}
\endgroup

For example, moving purple onto green from the practice state is legal and
produces [["yellow", "blue", "red"], ["green", "purple"]], but does not
finish the task. Moving red directly to the table from the initial state is
illegal because purple covers it. Replaying this illegal move returns
valid=false, \texttt{goal\_\allowbreak{}reached}=false, the unchanged initial state, and the reason
"illegal step 1: block 'red' is not on top of a stack".

The retained second draft supplies four demonstrations, each evaluated
independently from the original practice state:

\begin{center}
\small
\begin{tabular}{@{}lcc@{}}
\hline
Demonstration & Valid & Goal reached \\
\hline
purple $\to$ green & true & false \\
purple $\to$ table & true & false \\
purple $\to$ table (duplicate example) & true & false \\
purple $\to$ purple & false & false \\
\hline
\end{tabular}
\end{center}

All four received correct=true from the dynamics judge, giving score 1.0.
Both dynamics authors reached this score; author order selected author 0.
The duplicate table-move demonstration is present in the saved program and
does not add independent rule coverage.

The revision history also exposes a limitation of rendered demonstration
feedback. Author 0's first draft correctly rejected moving the covered red
block, but the judge treated red as the top block and marked that outcome
incorrect, producing score 0.75. This judgment is consistent with the
renderer cropping purple from the initial view. The second draft replaced
that demonstration with another purple-to-table move and received score
1.0. We therefore do not interpret the score increase as proof that a faulty
transition rule was repaired. It records improved agreement on the author's
chosen demonstrations; final plan correctness is checked independently.

\subsubsection{Planner and recorded execution}
\label{app:blocks-planner}

The selected planner implements solve(state, \texttt{time\_\allowbreak{}limit\_\allowbreak{}s}=20, seed=0).
Its fast constructive branch first moves blocks to the table until every
stack is a singleton, then builds the target stacks from bottom to top.
It tries goal-stack orderings and verifies the resulting arrangement using
the copied transition logic. Its source also contains a time-bounded A*
fallback; the constructive branch suffices for this example. The relevant
dispatch is:

\begingroup
\footnotesize
\begin{verbatim}
# Fast constructive solution.
flat_state, flatten_plan = flatten(initial)
if flat_state is not None:
    assembly = try_assembly(flat_state)
    if assembly is not None:
        return flatten_plan + assembly
\end{verbatim}
\endgroup

Planner construction used four symbolic practice states: the observed
practice reading and three distinct variants accepted by the selected
dynamics model. These variants are additional symbolic inputs, not additional
observed scenes. Each of the three planner authors achieved score 1.0 on
its first draft; author order selected author 0. For this selected program,
all four practice calls were valid and reached their goals. Recorded elapsed
times ranged from 0.118 to 0.128 seconds, including fresh worker setup, below
the 20-second allowance.

The saved plan for \texttt{blocks\_\allowbreak{}sim\_\allowbreak{}000} is:

\begingroup
\footnotesize
\begin{verbatim}
[
  {"block": "purple", "onto": "table"},
  {"block": "red",    "onto": "table"},
  {"block": "blue",   "onto": "table"},
  {"block": "yellow", "onto": "red"},
  {"block": "blue",   "onto": "yellow"},
  {"block": "purple", "onto": "green"}
]
\end{verbatim}
\endgroup

The following trace was reproduced by executing successive prefixes of that
saved plan with the unchanged selected dynamics model. Each bracketed stack
is ordered bottom to top; the order of stacks is immaterial.

\begin{center}
\footnotesize
\begin{tabular}{@{}rlp{0.64\linewidth}@{}}
\hline
Step & Action & Current stacks after the action \\
\hline
0 & Initial & \texttt{[yellow, blue, red, purple] [green]} \\
1 & purple $\to$ table & \texttt{[yellow, blue, red] [green] [purple]} \\
2 & red $\to$ table & \texttt{[yellow, blue] [green] [purple] [red]} \\
3 & blue $\to$ table & \texttt{[yellow] [green] [purple] [red] [blue]} \\
4 & yellow $\to$ red & \texttt{[green] [purple] [red, yellow] [blue]} \\
5 & blue $\to$ yellow & \texttt{[green] [purple] [red, yellow, blue]} \\
6 & purple $\to$ green & \texttt{[green, purple] [red, yellow, blue]} \\
\hline
\end{tabular}
\end{center}

Every move acts on an exposed block and uses a legal destination. The final
support relations are green-on-table, purple-on-green, red-on-table,
yellow-on-red, and blue-on-yellow. They exactly match the goal. This plan
satisfies the task's success criterion; the experiment does not require or
establish minimum plan length.

The historical evaluation checks plans against the reference initial and
goal stacks using the independent Blocks verifier after component selection.
It records "solved" for this instance and for all 19 non-practice instances:
20/20 overall and 19/19 outside the authoring practice scene. Reader selection
uses the unlabeled observation pool, so these 19 scenes are outside the
revision loop but are not a separate unseen reader-selection split. The
local replay used for this appendix reproduces the archived reading and
plan exactly, recomputes the final state, and passes the same independent
verifier without making new model calls.

The recorded 11 model calls in the planning-stage evaluation comprise
dynamics/planner construction and dynamics judging. They exclude the earlier
renderer and perceiver construction and should not be reported as the total
cost of constructing the complete system.



\end{document}